\documentclass[sigconf, screen, dvipsnames, prologue]{acmart}

\usepackage{multicol}
\usepackage{dsfont}
\usepackage{bbm}
\usepackage{amsmath}
\usepackage{colortbl}
\usepackage{xcolor}
\usepackage{ulem}
\usepackage{soul}
\definecolor{blue}{rgb}{0.1, 0.3, 0.7}    % 深蓝色
\definecolor{orange}{rgb}{1, 0.4, 0}    % 深橙色
\definecolor{yellow}{rgb}{1, 0.7, 0}     % 深黄色

\title{
Robust OOD Graph Learning via Mean Constraints and Noise Reduction
}
\author{Yang Zhou}
\affiliation{%
  \institution{Independent Researcher}
  \country{China}
}
\email{zz1999999999@outlook.com}

\author{Xiaoning Ren}
\affiliation{%
  \institution{University of Science and Technology of China}
  \city{Hefei}
  \country{China}
}
\email{hnurxn@mail.ustc.edu.cn}

\newtheorem{theorem}{Theorem}

\newtheorem{assumption}{Assumption}
\newtheorem{definition}{Definition}

\begin{document}

\begin{abstract}
% Graph Out-of-Distribution (OOD) problems in classification often result in significant performance degradation, especially when facing imbalanced graph categories and noisy graph structures. This paper addresses two key challenges in Graph OOD scenarios: (1) the poor performance of minority classes caused by category imbalance, and (2) the increased vulnerability of these minority classes as noise within graphs becomes more prevalent. To mitigate these issues, we propose two complementary methods. Specifically, we introduce the Constrained Mean Optimization (CMO), which enhances the model's performance of minority classes via using more similar instances in worst-case scenarios. Additionally, we design a Neighbor-Aware Noise Reweight (NNR), where each graph’s contribution is adjusted based on internal structural coherence, effectively reducing the impact of noise for training. Theoretical analysis supports the effectiveness of our approach, and experimental results on both synthetic and real-world graph datasets demonstrate notable gains in Graph OOD generalization and classification accuracy. 
Graph Out-of-Distribution (OOD) classification often suffers from sharp performance drops, particularly under category imbalance and structural noise. This work tackles two pressing challenges in this context: (1) the underperformance of minority classes due to skewed label distributions, and (2) their heightened sensitivity to structural noise in graph data. To address these problems, we propose two complementary solutions. First, Constrained Mean Optimization (CMO) improves minority class robustness by encouraging similarity-based instance aggregation under worst-case conditions. Second, the Neighbor-Aware Noise Reweighting (NNR) mechanism assigns dynamic weights to training samples based on local structural consistency, mitigating noise influence. We provide theoretical justification for our methods, and validate their effectiveness with extensive experiments on both synthetic and real-world datasets, showing significant improvements in Graph OOD generalization and classification accuracy.
The code for our method is available at: \url{https://anonymous.4open.science/r/CMO-NNR-2F30}.
\end{abstract}    
\maketitle
\section{Introduction}
\label{sec:intro}
% Graph-based learning has become an essential tool for various real-world applications, such as social networks, biological interactions, and recommendation systems, due to its ability to model complex, interconnected data. However, one significant challenge faced by graph-based models is the presence of OOD data. This issue arises particularly in real-world scenarios where the data is often imbalanced and noisy. Such data characteristics can severely degrade model performance and hinder its ability to generalize, especially when dealing with minority classes or groups. In these cases, models may struggle to capture critical patterns, leading to biased predictions. Addressing these issues is essential for ensuring the robustness and reliability of graph-based models in dynamic, real-world environments.
Graph-based learning has become a fundamental approach for modeling relational data in diverse domains, such as social networks, biological systems, and recommendation platforms\cite{dai2016deep,you2019hierarchical,hsu2021retagnn,yang2020multisage,wu2021recovering}. However, in practical applications, graphs collected from different environments often exhibit distribution shifts due to variations in structural patterns, feature distributions, or labeling rules. These shifts pose significant challenges to model generalization, leading to the emergence of Graph OOD research as a growing field focused on improving robustness across graph distribution shifts\cite{wu2024graph1,li2022out,liu2024beyond,wang2024distributionally}.

\begin{figure}[t]
  \centering
  \includegraphics[width=0.8\linewidth]{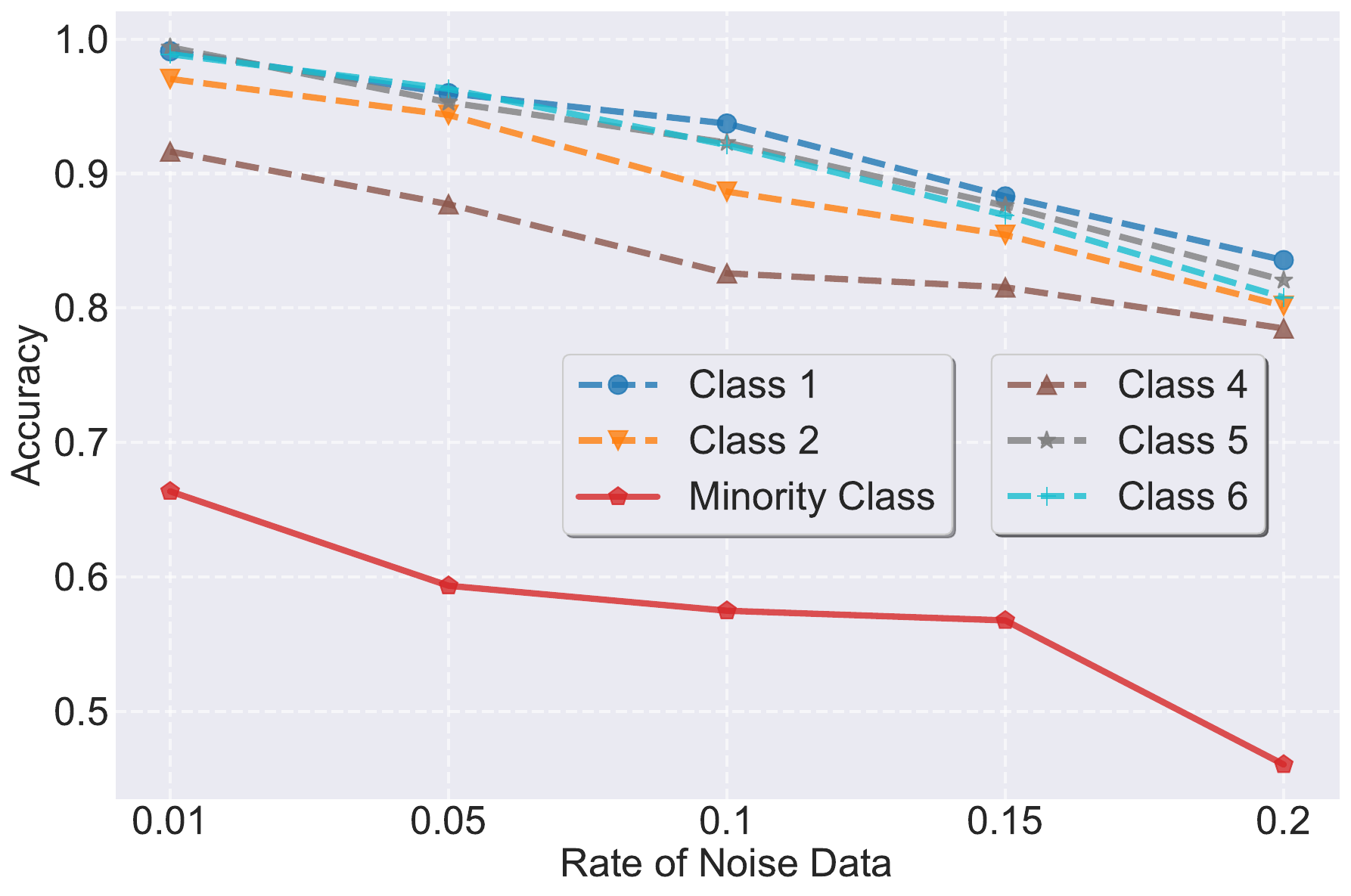}
  \caption{Accuracy for each category in different noise data rate}
  \label{fig:Noisedata0}
\end{figure}
Recently, there has been a rise in Graph OOD research\cite{Li2022OodGNN,wu2022DISCOVERING,Fan2022Debiasing,Wu2022HANDLING,Chen2023Does,Wu2024Graph,Jia2024Graph} that aim to improve generalization performance on complex distribution shifts. For instance, \textbf{GALA}\cite{Chen2023Does} use contrastive learning to maximize intra-class mutual information on different environments to learn causally invariant representations for graph. From another perspective, \textbf{DIR}\cite{wu2022DISCOVERING} and \textbf{IGM}\cite{Jia2024Graph} directly handle data shifts directly by employing Empirical Risk Minimization (\textbf{ERM}), which is implemented by designing new environments. 

In this work, we identified a Graph OOD issue that has been overlooked by existing research, i.e., \textit{the performance of minority classes decreases caused by imbalanced and noisy data}. As shown in Figure \ref{fig:Noisedata0}, the accuracy of the minority class is lower and will be exacerbated as noise increases.

% One important source of the OOD problem in graph learning is the imbalanced distribution of data, where certain classes contain significantly fewer samples than others. This imbalance leads to biased model training, as the learning process is dominated by the majority classes, causing the model to underperform on the minority classes. Consequently, models tend to prioritize optimizing performance for well-represented classes, while struggling to capture meaningful patterns from underrepresented ones. This issue is especially problematic in applications where minority classes are critical, such as detecting fraudulent transactions or identifying rare diseases in biomedical networks.
% Class imbalance is an important challenge in Graph OOD learning, where distribution shifts are amplified by the uneven sample sizes across classes. Under such imbalance, the model tends to overfit majority classes while failing to extract stable patterns from minority classes, which weakens its ability to generalize across environments. Noisy data is another factor that exacerbates the Graph OOD generalization, as noise cause a negative impact on generalization. Noise disrupts the extraction of consistent representations and disproportionately affects minority classes, whose limited samples are more sensitive to corrupted information. 
Class imbalance is an important challenge in Graph OOD learning, where distribution shifts are amplified by uneven sample sizes across classes. Under such imbalance, the model tends to overfit the majority classes while failing to extract stable patterns from minority classes, weakening its performance in minority class predictions\cite{anand1993improved,wei2022open}. Noisy data is another factor that exacerbates Graph OOD generalization, as noise negatively impacts model performance. It disrupts the extraction of consistent representations and disproportionately affects minority classes, whose limited samples are more sensitive to corrupted information, leading to further degradation\cite{cho2021data,gamberger1999experiments}.

To address these challenges, we propose two methods: CMO and NNR. CMO focuses on solving the class imbalance problem. It adds a constraint about the mean value of graph features during the update process, helping the model identify the worst-performing cases more effectively than traditional Distributionally Robust Optimization (DRO) methods. As a result, CMO balances the learning process and improves performance across all classes. NNR aims to reduce the impact of noisy data. During training, it assigns lower weights to samples likely to be noisy. To estimate the likelihood of each instance being noise, NNR counts the number of same-class neighbors within a certain distance. Fewer same-class neighbors suggest a higher probability of noise. By reducing the influence of these suspected noisy samples, especially in small classes, NNR helps the model focus on more reliable data. This improves generalization and enhances robustness against noise.

% To address these challenges, we propose two novel methods CMO and NNR. COM We observe that existing Graph OOD constraints are often applied in a fixed manner, without exploring more efficient ways to identify the worst-performing group. To address this, we introduce a prior constraint on the mean of the aggregated feature distance in Graph OOD environments, optimizing the update strategy of $q$, which represents the weight assigned to each sample during training to mitigate distribution shifts. Second, we mitigate the impact of noisy data by incorporating a weighted loss adjustment, where each data point's loss is modulated based on its similarity to other points, reducing the effect of noise on model optimization.

\noindent
Our contributions can be summarized as follows:

We propose a novel constrained optimization method for updating 
$q$, improving stability and adaptability in OOD settings.

We introduce a noise-aware loss weighting strategy that reduces the influence of noisy data, enhancing model generalization.

We provide a comprehensive theoretical analysis of both methods, demonstrating their effectiveness in addressing OOD challenges.

% We created several groups of synthetic datasets that simulate the real world to test our methods.

% We conduct extensive empirical evaluations on both synthetic and real-world datasets, validating the effectiveness of our proposed methods.
We design synthetic datasets that simulate real-world scenarios to analyze the impact of data imbalance and noise, and conduct comprehensive evaluations on both synthetic and real-world datasets\cite{2022arXiv220109637J} to verify the effectiveness of our proposed methods.

\section{Preliminaries}
We consider a graph classification problem in a cross-environment setting\cite{Chen2023Does}. 
\begin{assumption}
In each environment \( e \), every graph \( G_i \) belonging to class \( c \) is assumed to consist of two components, each of which is modeled as a Gaussian distribution:

\begin{itemize}
    \item \textbf{Invariant feature} \( x_{inv} \sim \mathcal{N}(\mu_c, \sigma^2 I) \), which reflects the inherent semantics associated with class \( c \). This feature is common across environments and captures the casual representation of the class.
    \item \textbf{Spurious feature} \( x_{spu} \sim \mathcal{N}(\mu_c^e, \sigma^2 I) \), which captures the environment-specific bias in \( e \).
\end{itemize}
Each class \( c \) is modeled by a Gaussian distribution with mean vector \( \mu_c \), and the environment-specific spurious feature has a Gaussian distribution with mean \( \mu_c^e \) for each environment \( e \).
\label{assumption0}
\end{assumption}

% \begin{assumption}
% The mean vectors \( \{\mu_c\} \) of different classes and the environment-specific means \( \{\mu_c^e\} \) are assumed to form orthonormal bases, i.e., for any pair of classes \( c \) and \( c' \), we have:
% \begin{equation}
% \langle \mu_c, \mu_{c'}^e \rangle = 0.
% \label{eq:orthonormality}
% \end{equation}
% \end{assumption}

As shown in ~\ref{fig:Noisedata0}, the performance of the minority class is often worse than that of the other classes. This poorer performance thus often causes a higher loss value, indicating that the model is struggling to make accurate predictions for these minority categories. Since loss is directly related to the model's prediction error, a higher loss suggests that the minority class is more likely to be classified as part of the “worst-case" scenarios during training.

DRO algorithms aim to identify and optimize for the worst-case scenarios\cite{sagawa2019distributionally}. These methods enhance the model's generalization ability by focusing on the most challenging cases, which often include high-loss instances. Given this, we propose applying DRO algorithms as foundation to address the performance gap in the minority class.
\subsection{DRO Algorithms}
Traditional DRO algorithms usually generate datasets that satisfy a constraint in the training data, and optimize the worst-performing data in order to improve the generalization ability of the model. Typically, the optimization objective is:
\begin{equation}
    \min_{f \in F} \{ R(f, q) \}
\end{equation}
where the objective function \( R(f, q) \) is defined as:
\begin{equation}\label{eq:TraditionalOOD}
    R(f, q) \equiv \max_{q \in \Delta_m} \sum_{i=0}^{n} q_i \mathbb{E}_{(x, y) \sim P(X, Y)} [\ell(f(x), y)]
\end{equation}
and the constraint is:
\begin{equation}
\text{s.t.} \ D(Q \| P) < \rho,
\end{equation}
\begin{equation}
Q = \left\{ \sum_{i=1}^n q_i P_i \mid q \in \Delta_m \right\}
\end{equation}
$Q$ represents the mixture of the test data distribution, \( P = \sum_{i=1}^n P_i \) is the training data distribution, and \( q_i \) is the weight of each group \( P_i \).

The constraint \( D(Q \| P) < \rho \) aims to ensure that the newly generated dataset maintains an appropriate difference from the original training data, thereby enhancing the model's generalization ability while avoiding significant deviation from the original data distribution.

\subsection{Divergence}
In this study, we focus on the f-divergences from the Cressie-Read family\cite{cressie1984multinomial}, defined by the parameter \( k \). Specifically, we define the function:
\begin{equation}
f_k(t) = k(k-1) t^{k} - k t + k - 1,
\end{equation}
and construct the divergence:
\begin{equation}
D_k(Q \| P) = \int f_k\left( \frac{dP}{dQ} \right) dP.
\end{equation}
When \( k = 1 \), this divergence reduces to the KL divergence. When \( k = 2 \), it becomes the Chi-squared divergence.

\begin{figure*}
   \centering
    \includegraphics[width=0.70\linewidth]{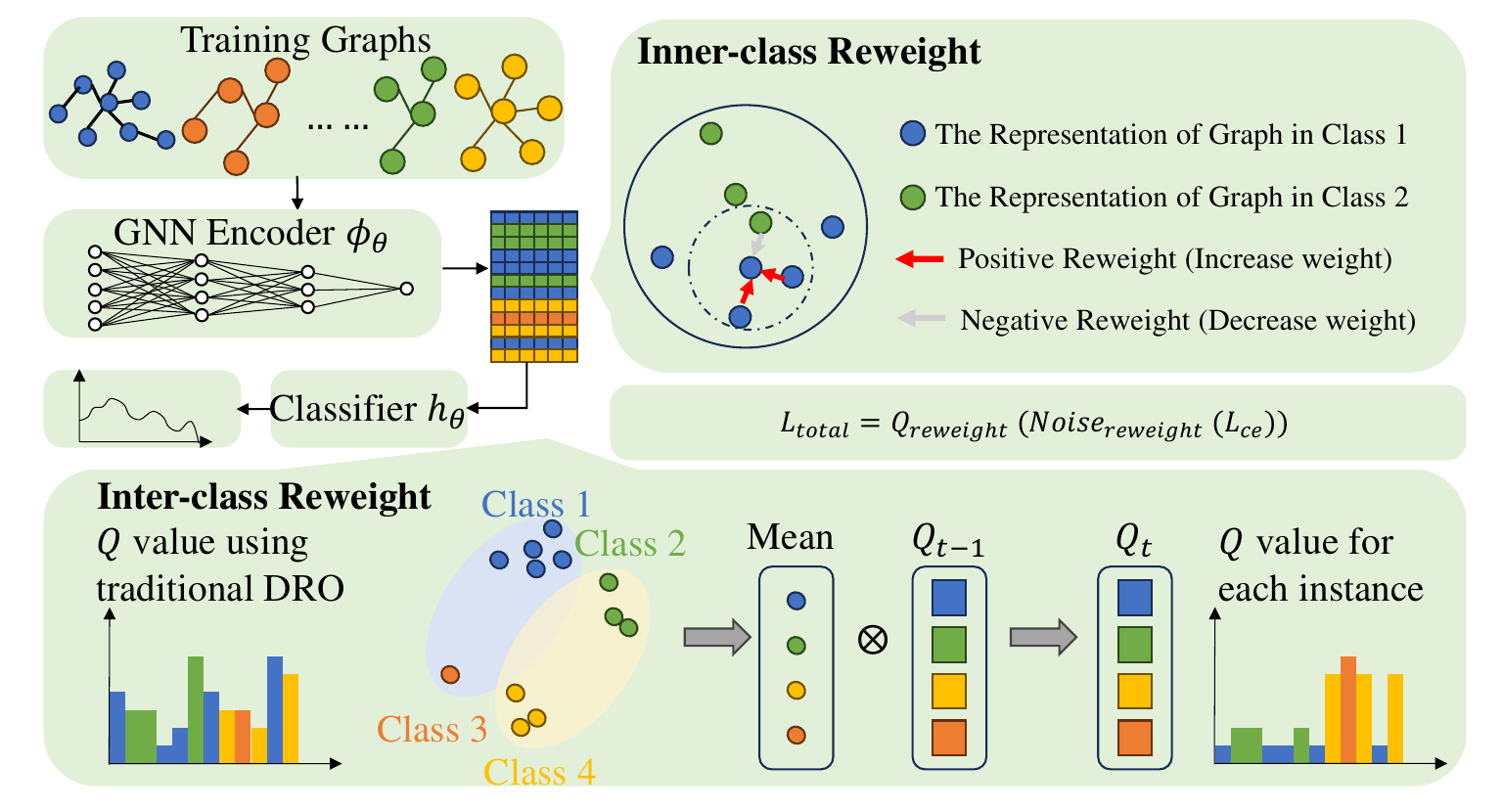}
    \caption{Our framework is composed of CMO and NNR, corresponding to inter-class and inner-class reweighting, respectively.}
    \label{fig:enter-label} 
\end{figure*}

\section{Methods}
In this section, we present two parallel methods (shown in ~\ref{fig:enter-label}) designed to address the poor performance of minority class due to data imbalance problem and the impact of noisy data in Graph Neural Networks (GNNs). 

% \begin{figure*}
%   \centering
%   \begin{subfigure}{0.68\linewidth}
%     \fbox{\rule{0pt}{2in} \rule{.9\linewidth}{0pt}}
%     \caption{An example of a subfigure.}
%     \label{fig:short-a}
%   \end{subfigure}
%   \hfill
%   \begin{subfigure}{0.28\linewidth}
%     \fbox{\rule{0pt}{2in} \rule{.9\linewidth}{0pt}}
%     \caption{Another example of a subfigure.}
%     \label{fig:short-b}
%   \end{subfigure}
%   \caption{Example of a short caption, which should be centered.}
%   \label{fig:short}
% \end{figure*}

\subsection{Constrained Mean Optimization (CMO)}
Traditional DRO algorithms typically maximize the loss function to find the worst-performing distribution \( Q \). However, this process ignores an important prior: for classification problems, smaller distributional differences between classes often lead to worse classification performance. For example, suppose we are classifying Alaskan Malamutes, Siberian Huskies, and cats. In this case, more worst-case scenarios are likely to occur between Alaskan Malamutes and Siberian Huskies, as these two breeds have more similar representations compared to cats. This class similarity is a crucial prior knowledge that can be used to improve the DRO algorithms.

To effective utilize the prior information of classification problems, we introduce a class-wise distribution similarity constraint, making the optimization objective:
\begin{equation}
\min_{f \in F} \{ R(f, q) \},
\end{equation}
where \( R(f, q) \) is still defined as:
\begin{equation}
R(f, q) \equiv \max_{q \in \Delta_m} \sum_{i=0}^{n} q_i \mathbb{E}_{(x, y) \sim P(X, Y)} [\ell(f(x), y)],
\end{equation}
and the updated constraint is:
\begin{equation}
\text{s.t.} \ D(Q \| P) < \rho_1, \Delta(Q) < \rho_2.
\end{equation}
\begin{equation}
\Delta(Q) = \sum_{i=1}^{n} \sum_{j=i+1}^{n} \left( \mu(Q_i) - \mu(Q_j) \right)^2.
\end{equation}
\(\Delta(Q)\) represents the sum of squares of the feature differences between classes.

To solve the optimization problem with these constraints, we use the Lagrange multiplier method to construct the Lagrangian dual function:
\begin{align}\label{eq:Lagrange}
        L = \sum_{i=0}^{n} q_i &\mathbb{E}_{(x, y) \sim P(X, Y)} [\ell(f(x), y)] \\& \qquad - \lambda_1 (D(Q \| P) - \rho_1) - \lambda_2 (\Delta(Q) - \rho_2).\nonumber
\end{align}
Let \( \theta \) represent the parameters of the model \( f(x) \), and the update formulas for \( \theta \) and \( q \) are:
\begin{equation}
q^{(t+1)} = q^{(t)} + \eta_q \nabla_q L, \quad \theta^{(t+1)} = \theta^{(t)} - \eta_\theta \nabla_\theta L,
\end{equation}
where \( \eta_q \) and \( \eta_\theta \) are the update step sizes for \( q \) and \( \theta \), respectively.

\subsection{Convergence Analysis}
In this section, we prove that by choosing appropriate learning rates \( \eta_\theta \) and \( \eta_q \), the proposed optimization method can ensure the convergence of the parameters \( \theta \) and \( q \) for both convex and non-convex loss functions.

\begin{definition}\label{De:Lipschitz_Continuity}
A mapping \( f: X \to \mathbb{R}^m \) is \( L \)-Lipschitz continuous if for any \( x, y \in X \),
\begin{equation}
\| f(x) - f(y) \| \leq L \| x - y \|.
\end{equation}
\end{definition}
\begin{definition}\label{De:Smoothness}
A function \( f: X \to \mathbb{R} \) is \( L \)-smooth if it is differentiable on \( X \) and the gradient \( \nabla f \) is \( G \)-Lipschitz continuous, i.e., 
\begin{equation}
\| \nabla f(x) - \nabla f(y) \| \leq G \| x - y \| \quad \text{for all } x, y \in X.
\end{equation}
\end{definition}

\begin{assumption}\label{assumption1}  
We assume that the loss function satisfies the above two properties (~\ref{De:Lipschitz_Continuity} and ~\ref{De:Smoothness}).  
\end{assumption}

\textbf{Convex Loss Functions}
For convex loss functions, we estimate the convergence rate based on the expected number of stochastic gradient updates. To achieve a duality gap of \( \epsilon \)\cite{nemirovski2009robust}, the optimal convergence rate for solving stochastic min-max problems \( O(1/\epsilon^2) \) assuming the problem is convex-concave. The duality gap for the pair \( (\tilde{\theta}, \tilde{q}) \) is
\begin{equation}
\max_{q \in \Delta_m} R(\tilde{\theta}, q) - \min_{\theta \in \Theta} R(\theta, \tilde{q}).
\end{equation}
In cases of strong duality, the pair \( (\tilde{\theta}, \tilde{q}) \) is considered optimal when the duality gap is zero. Our method is shown to achieve the optimal rate of \( O(1/\epsilon^2) \) in this setting.
Consider the problem in ~\ref{eq:Lagrange}.
\begin{theorem}\label{Th:theorem1}
Assuming the loss function is convex and Assumption \ref{assumption1} is satisfied. Let \( \Theta \) be bounded by \( R_{\theta} \), with \( \mathbb{E} \left[ \| \nabla_{\theta} R(\theta, q) \|_2^2 \right] \leq \hat{G}_{\theta}^2 \) and \( \mathbb{E} \left[ \| \nabla_{q} R(\theta, q) \|_2^2 \right] \leq \hat{G}_{q}^2 \), our method satisfies:
\begin{equation}
    \begin{aligned}
        & \mathbb{E} \left[ \max_{q \in \Delta_m} R(\theta^T, q) - \min_{\theta \in \Theta} R(\theta, q^T) \right] \leq 2 G_q  + 2 \| R_{\theta} \|_2 G_{\theta} \\&+ \frac{1}{\sqrt{T}} \left( \mathbb{E} \left[ \| \theta^1 - \theta \|_2 \right] G_{\theta} \right)  + \frac{1}{\sqrt{T}} \left( \mathbb{E} \left[ \| q^1 - q \|_2 \right] G_q \right)
    \end{aligned}
\end{equation}
\end{theorem}

% ~\ref{sec:method1}
The proof of ~\ref{Th:theorem1} is detailed in Supplementary material. This inequality shows that the expectation of the dual gap monotonically decreases as the number of iterations \( T \) increases, converging at a rate of \( O(1/T) \), thus ensuring the convergence of the optimization process under convex loss functions.

\textbf{ Non-Convex Loss Functions}
For non-convex loss.
\begin{theorem}\label{Th:theorem2}
Based on Assumption \ref{assumption1}
\begin{equation}
\lim_{T \to \infty} \sum_{t=1}^{T} \mathbb{E} \left[ \| \nabla_{\theta} L(\theta_t, q_t) \| \right] = 0.
\end{equation}
\end{theorem}

% ~\ref{sec:method2}
The proof of~\ref{Th:theorem2} is detailed in Supplementary material. This result shows that as the number of iterations increases, the cumulative gradient norm tends to zero, ensuring that in the case of non-convex losses, the algorithm can find a stationary point, guaranteeing overall convergence.

\subsection{Neighbor-Aware Noise Reweighting (NNR)}
% In DRO algorithms, the model typically tries to optimize the worst-performing cases, which results in overemphasis on noisy data. Noisy data often manifests as poor performance in the ideal model, and traditional OOD methods, as described in ~\ref{eq:TraditionalOOD}, prefer to utilize poor performed data during training,  thus interfering with the training process and causing the model to deviate from the true target. To address this issue, we propose a neighbor-based weight adjustment method. The core idea is as follows:
In DRO algorithms, the model typically focuses on optimizing the worst-performing cases, which often leads to overemphasis on noisy data. \textbf{This can cause a significant issue, as noisy data tends to negatively impact the model's performance by introducing misleading patterns.} Noisy data usually manifests as poor performance in the ideal model, and traditional Graph OOD methods, as described in ~\ref{eq:TraditionalOOD}, tend to prioritize such poorly performing data during training. This causes disturbance in the training process, leading the model to deviate from the true target. \textbf{To resolve this issue, we propose a neighbor-based weight adjustment method}, which aims to reduce the influence of noisy data by adjusting the weight based on data similarity.

% In DRO algorithms, the model typically focuses on optimizing the worst-performing cases, often overemphasizing noisy data. \textbf{This can cause a significant issue, as noisy data negatively impacts the model's performance by introducing misleading patterns.} Noisy data usually manifests as poor performance in the ideal model, and traditional Graph OOD methods, as described in ~\ref{eq:TraditionalOOD}, tend to prioritize such poorly performing data during training. This disturbs the training process, leading the model to deviate from the true target. \textbf{To resolve this, we propose a neighbor-based weight adjustment method}, reducing the influence of noisy data by adjusting the weight based on data similarity.

 % For each data sample, the reliability of these data is measured by counting the number of "same" data in its local neighborhood. Assuming that the data in each class follows a Gaussian distribution, based on prior knowledge, when there are more same instances surrounding a sample, it is more likely to be real data and should be assigned a higher weight. Conversely, when there are fewer similar instances, it is more likely to be noisy data and should be assigned a lower weight.

\textbf{Basic Idea:} According to Assumption~\ref{assumption0}, since the invariant features of class \( c \) follow a Gaussian distribution \( x_{inv} \sim \mathcal{N}(\mu_c, \sigma^2 I) \), a sample with more neighbors sharing similar labels is more likely to be reliable and consistent with the true distribution, thus assigned a higher weight. Conversely, if there are fewer such neighboring instances, the sample is considered more likely to be noisy and is assigned a lower weight.

\textbf{Normalization:} Since imbalanced data distribution is very common in the Graph OOD problem, it is necessary to normalize the number of the same data in the neighborhood to eliminate the impact of different class sizes.

\textbf{Implementation Details:} 
First, for each graph data \(G_i\), we calculate the “same" neighbors within a certain distance, denoted as \(N_{hm}(G_i)\). Then, for each sample \(G_i\), we calculate the total number of graph instances in training progress, denoted as \(N_{sum}(G_i)\). The weight \(w_i\) of data \(G_i\) is defined as:
\begin{equation}
w_i = \frac{N_{hm}(G_i)}{N_{sum}(G_i)}.
\end{equation}

And, this weight \(w_i\) is applied to the loss calculation for graph \(G_i\), resulting in the weighted loss function:
\begin{equation}
\ell_i = w_i \ell_i^{raw}.
\end{equation}

This method helps to reduce the contribution of noisy data, thereby improving the robustness of the model.

\subsection{OOD Generalization Error Bound}
% \subsubsection{Data Model and Definitions}
% We consider a graph classification problem in a cross-environment setting. In each environment \( e \), every graph \( G_i \) belonging to class \( c \) is assumed to consist of two components:

% \begin{itemize}
%     \item \textbf{Invariant feature} \( x_{inv} \sim \mathcal{N}(\mu_c, \sigma^2 I) \), which reflects the inherent semantics associated with class \( c \).
%     \item \textbf{Spurious feature} \( x_{spu} \sim \mathcal{N}(\mu_c^e, \sigma^2 I) \), capturing the environment-specific bias in \( e \).
% \end{itemize}

% We assume that the mean vectors \( \{\mu_c\} \) of different classes and the environment-specific means \( \{\mu_c^e\} \) form orthonormal bases. That is, for any \( c, c' \), we have:
% \begin{equation}
% \langle \mu_c, \mu_{c'}^e \rangle = 0.
% \label{eq:orthonormality}
% \end{equation}
% Additionally, we assume that \( x_{inv} \) and \( x_{spu} \) are independent and both follow Gaussian distributions with the same variance \( \sigma^2 \).
In this section, we derive an OOD generalization error bound to demonstrate that NNR effectively minimizes OOD error. To achieve this, we adopt a PAC-Bayesian framework, following the approach proposed by Ma et al. (2021)\cite{ma2021subgroup}. The total set of graphs in the environment \( e \) is denoted by \( G^e \), with training and test sets represented as \( G_{tra}^e \) and \( G_{te}^e \), respectively.

% \begin{definition}[Local Homophily-based Neighbor Count]
% For a given graph \( G_i \), we define the local homophily-based neighbor count \( N^{hm}(G_i) \) as the number of same-class graphs within a feature-space distance threshold \( \tau \), which measures the density of similar local samples.
% \end{definition}

\subsubsection{Graph Embedding and Global Features}
\begin{definition}[Graph Embedding]
For each graph \( G_i \), we obtain an embedding \( g_i \in \mathbb{R}^d \) via global pooling (e.g., Readout) from a GNN. The classification function \( f: \mathbb{R}^d \to \mathbb{R}^c \) satisfies the Lipschitz continuity condition:
\begin{equation}
\|f(g_i) - f(g_j)\| \leq C \| g_i - g_j \|,
\label{eq:lipschitz}
\end{equation}
where \( L \) is the Lipschitz constant, ensuring that nearby embeddings yield similar classification outputs.
\end{definition}

\begin{definition}[Feature Distance Aggregation]
To characterize the difficulty of cross-environment generalization, we define the graph embedding distance between the test environment \( G_{te}^e \) and the training environment \( G_{tra}^e \) as:
\begin{equation}
\Gamma = \max_{G_j \in G_{te}^e} \min_{G_i \in G_{tra}^e} \| g_i - g_j \|_2.
\label{eq:gamma_definition}
\end{equation}
A larger \( \Gamma \) indicates that test graphs are farther from training graphs in the embedding space, which increases the generalization difficulty.
\end{definition}
\subsubsection{Margin Loss and Classifier Constraints}
\begin{definition}[Margin Loss]
The graph classification margin loss function for environment \( e \) is defined as:
\begin{equation}
L_{\gamma}^e(h) = \frac{1}{|G^e|} \sum_{G_i \in G^e} \mathbb{I} \left[ h(g_i)[y_i] \leq \gamma + \max_{c \neq y_i} h(g_i)[c] \right],
\label{eq:margin_loss}
\end{equation}
where \( h: \mathbb{R}^d \to \mathbb{R}^c \) is the classifier, \( \gamma \) is the margin threshold, and \( \mathbb{I} \) is the indicator function.
\end{definition}
% \begin{assumption}[Classifier Parameter Constraints]
% Let the classifier parameters be \( W_1, \dots, W_L \), and define the spectral norm upper bound as:
% \begin{equation}
% T = \max_{1 \leq l \leq L} \| W_l \|_2.
% \label{eq:spectral_norm}
% \end{equation}
% \end{assumption}
\begin{assumption}[Equal-Sized and Disjoint Near Sets]
For any \( 0 < m \leq M \), assume the near sets of each graph \( G_i \in G_{tra} \) with respect to \( G_{te} \) are disjoint and have the same size \( s_m \in \mathbb{N}^+ \).
\label{assumption2}
\end{assumption}

\begin{assumption}[Cross-Environment Embedding Coverage]
For any test environment \( e_{test} \), there exists a coverage radius \( \Gamma \) such that for every training graph \( G_i \in G_{tra}^e \), the neighboring test graph set
\begin{equation}
N_{\Gamma}(G_i) = \{ G_j \in G_{te}^e \mid \| g_i - g_j \|_2 \leq \Gamma \}
\end{equation}
is non-empty. \(g_i\) denotes as a Readout function. Additionally, we assume: There exists a constant \( s_e > 0 \) such that for most training graphs, \( |N_{\Gamma}(G_i)| \geq s_e \). The neighboring sets \( N_{\Gamma}(G_i) \) should exhibit minimal overlap to better characterize training sample coverage.
\label{assumption3}
\end{assumption}

\begin{assumption}[Concentrated Expected Loss Difference]
Let \( P \) be a distribution over the hypothesis space \( H \), defined by sampling the vectorized MLP parameters from \( \mathcal{N}(0, \sigma^2 I) \) for \(\sigma^2 \leq \frac{\left( \frac{\gamma}{8 \Gamma} \right)^{\frac{2}{L}}}{2b \left( \lambda |G_{tra}| + \ln 2bL \right)}
\) For any \( L \)-layer GNN classifier \( h \in H \) with model parameters \( W_1^h, \dots, W_L^h \), define \(T_h := \max_{l=1,\dots,L} \| W_l^h \|_2\). Assume that there exists some \( 0 < \alpha < \frac{1}{4} \) such that
% \begin{equation}
% \begin{aligned}
%    \mathbb{P}_{h \sim P} \left( L_{\gamma/4}^{e_{test}}(h) - L_{\gamma/2}^{e_{train}}(h)  > |G_{tra}|^{-\alpha} + C\Gamma T \,\Big|\, T_h^L \Gamma > \frac{\gamma}{8} \right)\\ \leq \exp(-|G_{te}|^{2\alpha}). 
% \end{aligned}
% \end{equation}

\begin{equation}
\mathbb{P}_{h \sim P} \left( A > B \,\Big|\, T_h^L \Gamma > \frac{\gamma}{8} \right) \leq \exp(-|G_{te}|^{2\alpha}).
\end{equation}
\[
A = L_{\gamma/4}^{e_{test}}(h) - L_{\gamma/2}^{e_{train}}(h) \quad \text{and} \quad B = |G_{tra}|^{-\alpha} + C\Gamma T
\]
\label{assumption4}
\end{assumption}
\begin{theorem}
\label{Th:theorem3}
For any classifier \( h \in H \) and environment \( e^{test}\), under the Assumption \ref{assumption2} \ref{assumption3} \ref{assumption4}, when the number of training graphs \( |G_{tra}| \) is sufficiently large, there exists a constant \( \alpha \in (0, 1/4) \) such that with probability at least \( 1 - \delta \), the following holds:
\begin{align}
\label{eq:overllbound}
   & L^{(e^{test})} \leq L^{(e^{train})} + O \left( \frac{1}{|G_{tra}^e|} \sum_{i \in G_{tra}^e} \frac{1}{|G_{te}^e|} \right. \notag \\
    &\qquad\quad \left. \sum_{j \in G_{te}^e }  \sum_{c=1}^C\sum_{c' \neq c} \left[ \text{Term}_1 + \text{Term}_2 + \text{Term}_3 \right] \right)
\end{align}
\begin{equation}
    \text{Term}_1 = \frac{1}{\sigma^2} \left( (g_j^T w_j - g_i^T w_i)(\mu_{c'} - \mu_c) \right)
\end{equation}
\begin{equation}
    \text{Term}_2 = \frac{1}{2\sigma^2} \left( w_j^2 \left( f(\mu^{test}) \right) - w_i^2 \left( f(\mu^{train}) \right) \right)
\end{equation}
\begin{equation}
f(\mu) = \|\mu_{c}\|_2^2 - \|\mu_{c'}\|_2^2
\end{equation}
\begin{equation}
    \text{Term}_3 = \frac{1}{2\sigma^2} \left( (w_j^2 - w_i^2) \left( \|\mu_c\|_2^2 - \|\mu_{c'}\|_2^2 \right) \right)
\end{equation}
\end{theorem}
% ~\ref{sec:method3}
The detail of Theorem ~\ref{Th:theorem3} can be found in Supplementary material.  \( \mu_c \) and \( \mu_{c'} \) denote the class centers of classes \( c \) and \( c' \), respectively. \( g_i \) and \( g_j \) are the readout values of samples \( i \) and \( j \) from the training set \( G_{tra}^e \) and test set \( G_{te}^e \), respectively. \( w_i \) and \( w_j \) are the corresponding weights of samples \( g_i \) and \( g_j \). \( |G_{tra}^e| \) and \( |G_{te}^e| \) represent the sizes of the training and test sets for environment \( e \), and \( C \) is the total number of classes.
Based on the above ~\ref{eq:overllbound}, the error bound can be decomposed into three distinct terms.
\begin{table*}[t]
\centering
\resizebox{\textwidth}{!}{  % 自动调整宽度以适应页面
\begin{tabular}{@{}lcc|cc|cc@{}}
\toprule
Method & \multicolumn{2}{c}{Minority Class Accuracy (Average\%)} & \multicolumn{2}{c}{Minority Class Accuracy (Max\%)} & \multicolumn{2}{c}{All Classes Average Accuracy (Average\%)} \\ \midrule
 & \multicolumn{1}{c}{0.15} & \multicolumn{1}{c}{0.2} & \multicolumn{1}{c}{0.15} & \multicolumn{1}{c}{0.2} & \multicolumn{1}{c}{0.15} & \multicolumn{1}{c}{0.2} \\
\midrule
ERM & 54.9 $\pm$0.1 & 53.6 $\pm$0.4 & 60.3 & 63.1 & 80.9  & 76.7  \\
ERM+NNR & \textcolor{blue}{\underline{70.1}} $\pm$0.5 & 58.6 $\pm$0.6 & 83.6 & 73.4 & \textcolor{yellow}{\underline{\underline{81.1}}}  & \textcolor{blue}{\underline{77.0}}  \\
ERM+CMO-KL & \textcolor{yellow}{\underline{\underline{67.8}}} $\pm$0.5 & \textcolor{blue}{\underline{62.8}} $\pm$0.5 & 81.5 & 71.6 & 80.9  & \textcolor{yellow}{\underline{\underline{76.7}}}  \\
ERM+CMO-Chi & 67.5 $\pm$0.6 & \textcolor{orange}{\dotuline{61.6}} $\pm$0.5 & \textcolor{yellow}{\underline{\underline{85.0}}} & \textcolor{yellow}{\underline{\underline{74.7}}} & 81.0  & 76.7  \\
ERM+NNR+CMO-KL & \textcolor{orange}{\dotuline{67.9}} $\pm$0.8 & \textcolor{yellow}{\underline{\underline{59.3}}} $\pm$0.8 & \textcolor{blue}{\underline{85.6}} & \textcolor{blue}{\underline{80.2}} & \textcolor{blue}{\underline{81.1}}  & 75.5  \\
ERM+NNR+CMO-Chi & 67.2 $\pm$0.3 & 58.5 $\pm$0.7 & \textcolor{orange}{\dotuline{85.0}} & \textcolor{orange}{\dotuline{79.8}} & \textcolor{orange}{\dotuline{81.1}}  & 75.8  \\
CVaR & 55.0 $\pm$0.1 & 54.0 $\pm$0.2 & 60.3 & 62.7 & 81.0  & \textcolor{orange}{\dotuline{76.8}}  \\
Chisq & 57.3 $\pm$0.4 & 54.4 $\pm$0.3 & 72.3 & 62.8 & 81.1  & 76.7  \\
CVaR\_doro & 55.1 $\pm$0.2 & 52.0 $\pm$0.3 & 64.8 & 59.8 & 80.9  & 76.4  \\
Chisq\_doro & 55.6 $\pm$0.2 & 53.0 $\pm$0.3 & 67.7 & 61.3 & 80.8  & 76.6  \\
CVaR\_group & 55.5 $\pm$0.2 & 53.0 $\pm$0.2 & 69.9 & 62.3 & 80.9  & 76.7  \\
Variant & 53.0 $\pm$0.6 & 53.7 $\pm$1.0 & 67.7 & 69.0 & 78.1  & 74.0  \\
Group & 53.9 $\pm$0.2 & 52.4 $\pm$0.3 & 61.4 & 63.1 & 80.9  & 76.5  \\
Gradient & 56.6 $\pm$0.1 & 52.0 $\pm$0.3 & 63.8 & 60.2 & 81.0  & 76.4  \\
\bottomrule
\end{tabular}
}
\caption{Performance Results for Various Methods on Synthetic Datasets (In the table above, for the columns 'Minority Class Accuracy (Average\%)' and 'All Classes Average Accuracy (Average\%)', entries without a variance annotation indicate a variance of 0.0\%. The blue single underline denotes the first place, the orange dotted underline indicates the second place, and the yellow double underline represents the third place.)}
\label{tab:NR}
\end{table*}

\begin{enumerate}
    \item \textbf{Class Center Differences and Weighted Distance}: Measures the difference between class centers \( \|\mu_{c'} - \mu_c\|_2 \) and the weighted distance \( \| g_j^T w_j - g_i^T w_i \|_2 \) between samples, with \( w_i \) and \( w_j \) as sample weights, combining class centrality and sample distances.
    \item \textbf{Cross-Environment Feature Mean Differences}: Calculates the differences in feature means \( \|\mu_{c^{test}}\|_2^2 - \|\mu_{c'^{test}}\|_2^2 \) between the training and test environments, highlighting discrepancies in feature distributions.
    \item \textbf{Weighted Distribution of Different-Class Neighbor Ratios}: Assesses the weighted differences in feature distributions of different classes based on different-class neighbor ratios across environments.
\end{enumerate}

% \begin{enumerate}
%     \item \textbf{Class Center Differences and Weighted Distance}: This term measures the difference between class centers \( \|\mu_{c'} - \mu_c\|_2 \), and incorporates the weighted distance \( \| g_j^T w_j - g_i^T w_i \|_2 \) between samples \( g_i \) and \( g_j \) from the training and test sets, with \( w_i \) and \( w_j \) being the respective sample weights. It combines class centrality and sample distances.
%     \item \textbf{Cross-Environment Pseudo-Feature Mean Differences}: This term calculates the differences in pseudo-feature means \( \mu_c^{e} \) between the training and test environments, such as \( \|\mu_{c^{test}}\|_2^2 - \|\mu_{c'^{test}}\|_2^2 \) and \( \|\mu_{c^{train}}\|_2^2 - \|\mu_{c'^{train}}\|_2^2 \), highlighting discrepancies in feature distributions and potential misalignments.

%     \item \textbf{Weighted Distribution of Same-Class Neighbor Ratios}: This term assesses the weighted distribution differences of same-class neighbor ratios across environments, capturing how node relationships evolve and are influenced by training conditions.
% \end{enumerate}
\noindent
Based on the formulas, we can indirectly adjust the weight \( w \) through the parameter \( \Gamma \), thereby controlling the model's Generalization Error Bound.
 % ~\ref{sec:SE}
\section{Experiment}
In this section, we provide a detailed explanation of the experimental setup and present a subset of the results (due to space limitations). Additional experimental results can be found in the Supplementary material.

\subsection{Dataset}
To evaluate the effectiveness of the proposed method, we conduct experiments in both synthetic datasets and real-world datasets\cite{2022arXiv220109637J}.

\textbf{Synthetic Datasets}  
We have constructed a synthetic dataset consisting of two parts: invariant features, which determine the true label, and spurious features. To simulate real-world noise, we introduce two key parameters:
\begin{enumerate}
    \item \textbf{Noise Ratio} (\( \alpha \)): The probability that invariant features do not match the true label, with a proportion of samples having their labels randomly assigned. The noise ratio varies across {0.01, 0.05, 0.1, 0.15, 0.2}.
    \item \textbf{Correlation Strength} (\( \beta \)): The relationship between invariant and spurious features. When \( \beta = 1 \), spurious features are fully determined by the invariant features.
\end{enumerate}
The dataset, which we have synthesized, contains six classes: classes 1, 2, 4, 5, and 6 each with 3000 samples, and class 3 with 300 samples. Each class has 1000 samples for validation and testing. Various combinations of noise ratios and correlation strengths produce different dataset variations, which we use to test our model.

\textbf{Real-world Dataset}
In addition to the synthetic dataset, we also conducted experiments on real-world datasets\cite{2022arXiv220109637J} to verify the generalization ability of our method. The information for these three datasets is shown in ~\ref{tab:Realdata}.
\begin{table}[h]
\resizebox{0.45\textwidth}{!}{  % 自动调整宽度以适应页面
\begin{tabular}{@{}lcc|cc|cc@{}}
\toprule
 & \multicolumn{2}{c}{Train}
&\multicolumn{2}{c}{Validation} & \multicolumn{2}{c}{Test} \\
\midrule
& \multicolumn{1}{c}{Class 1} & \multicolumn{1}{c}{Class 2} & \multicolumn{1}{c}{Class 1} & \multicolumn{1}{c}{Class 2} & \multicolumn{1}{c}{Class 1} & \multicolumn{1}{c}{Class 2} \\
\midrule
Assay & 68 & 8325 & 67 & 4564 & 305 & 4338 \\
Scaffold & 82 & 8399 & 76 & 4568 & 277 & 4194 \\
Size & 38 & 5285 & 76 & 4587 & 289 & 4379 \\
\bottomrule
\end{tabular}
}
\caption{Information about the real-world dataset for training, validation, and testing for drugood\_lbap\_core\_ki.}
\label{tab:Realdata}
\end{table}

\begin{table*}[t]
\centering
\resizebox{\textwidth}{!}{  % 自动调整宽度以适应页面
\begin{tabular}{@{}lcc|cc|cc@{}}
\toprule
        Method & \multicolumn{2}{c}{drugood\_lbap\_core\_ki\_assay(Average\%)} & \multicolumn{2}{c}{drugood\_lbap\_core\_ki\_scaffold(Average\%)} & \multicolumn{2}{c}{drugood\_lbap\_core\_ki\_size(Average\%)}  \\ \midrule
        % \cmidrule(lr){2-3} \cmidrule(lr){4-5} \cmidrule(lr){6-7}
        & \multicolumn{1}{c}{Class 1} & \multicolumn{1}{c}{Class 2} & \multicolumn{1}{c}{Class 1} & \multicolumn{1}{c}{Class 2} & \multicolumn{1}{c}{Class 1} & \multicolumn{1}{c}{Class 2} \\
        \midrule
        ERM & 13.0 $\pm$0.4 & \textcolor{yellow}{\underline{\underline{96.2}}}  & 6.3 $\pm$0.1 & \textcolor{yellow}{\underline{\underline{97.6}}}  & 14.9 $\pm$0.6 & 97.2  \\
        ERM+NNR & 12.9 $\pm$0.3 & 95.7  & 6.9 $\pm$0.2 & 96.1  & 13.1 $\pm$0.6 & \textcolor{blue}{\underline{97.8}}  \\
        ERM+CMO-KL & \textcolor{yellow}{\underline{\underline{19.3}}} $\pm$0.7 & 90.6 $\pm$0.2 & 8.5 $\pm$0.4 & 96.1 $\pm$0.1 & 19.9 $\pm$0.9 & 94.6 $\pm$0.1 \\
        ERM+CMO-Chi & 19.2 $\pm$1.0 & 90.7 $\pm$0.3 & 10.1 $\pm$0.4 & 95.2 $\pm$0.1 & 19.1 $\pm$0.8 & 94.5 $\pm$0.1 \\
        ERM+NNR+CMO-KL & \textcolor{blue}{\underline{21.2}} $\pm$0.6 & 89.7 $\pm$0.3 & \textcolor{blue}{\underline{15.2}} $\pm$0.7 & 92.6 $\pm$0.3 & \textcolor{blue}{\underline{27.4}} $\pm$0.8 & 91.3 $\pm$0.2 \\
        ERM+NNR+CMO-Chi & \textcolor{orange}{\dotuline{21.2}} $\pm$0.9 & 90.6 $\pm$0.2 & \textcolor{orange}{\dotuline{14.7}} $\pm$0.7 & 92.1 $\pm$0.3 & \textcolor{orange}{\dotuline{25.6}} $\pm$1.0 & 91.3 $\pm$0.2 \\
        CVaR & 14.2 $\pm$0.4 & 94.3 $\pm$0.1 & 8.2 $\pm$0.7 & 95.4 $\pm$0.3 & 16.0 $\pm$0.4 & 96.7  \\
        Chisq & 12.1 $\pm$0.2 & 95.8  & 9.5 $\pm$0.5 & 95.0 $\pm$0.2 & 13.6 $\pm$0.7 & 97.0  \\
        CVaR\_doro & 9.6 $\pm$0.2 & \textcolor{orange}{\dotuline{96.7}}  & 5.9 $\pm$0.1 & 96.7  & 14.5 $\pm$0.3 & 96.1  \\
        Chisq\_doro & 13.0 $\pm$0.1 & 95.3 $\pm$0.1 & \textcolor{yellow}{\underline{\underline{12.7}}} $\pm$0.3 & 95.1 $\pm$0.1 & 15.0 $\pm$0.4 & \textcolor{yellow}{\underline{\underline{97.4}}}  \\
        CVaR\_group & 12.7 $\pm$0.3 & 95.8 $\pm$0.1 & 9.3 $\pm$0.2 & 96.0 $\pm$0.1 & \textcolor{yellow}{\underline{\underline{20.3}}} $\pm$1.2 & 96.4 $\pm$0.1 \\
        Variant & 15.5 $\pm$0.1 & 94.9 $\pm$0.1 & 6.3 $\pm$0.2 & \textcolor{orange}{\dotuline{97.7}}  & 17.2 $\pm$0.4 & \textcolor{orange}{\dotuline{97.7}}  \\
        Group & 14.3 $\pm$0.4 & 95.3 $\pm$0.1 & 8.9 $\pm$0.4 & 96.3 $\pm$0.2 & 12.9 $\pm$0.5 & 96.5  \\
        Gradient & 3.2 $\pm$0.2 & \textcolor{blue}{\underline{98.6}} $\pm$0.1 & 2.6 $\pm$0.1 & \textcolor{blue}{\underline{99.0}}  & 14.3 $\pm$1.0 & 95.7 $\pm$0.1 \\
\bottomrule
\end{tabular}
}
\caption{Performance Results for Various Methods in Real-World Datasets (In the table above, entries without a variance annotation indicate a variance of 0.0\%. The blue single underline denotes the first place, the orange dotted underline indicates the second place, and the yellow double underline represents the third place.)}
\label{tab:Realdata_Performance1}
\end{table*}

\subsection{Baseline Methods}

For our experiments, we selected a simple 3-layer Graph Neural Network (GNN) as the backbone model. As for the baseline methods, we chose the most commonly used ERM approach along with several other advanced techniques. These include:

\begin{itemize}
\item \textbf{ERM}\cite{pfau2024engineering}: Standard empirical risk minimization on training data.
\item \textbf{CVaR}\cite{zhai2021doro}: DRO method minimizing loss over the worst \(\alpha\) fraction of samples to improve tail performance.
\item \textbf{Chisq}\cite{zhai2021doro}: DRO method using \(\chi^2\)-divergence to reweight samples, emphasizing high-loss cases.
\item \textbf{CVaR\_doro}\cite{zhai2021doro}: CVaR extension with outlier robustness by filtering the top \(\epsilon\) high-loss samples.
\item \textbf{Chisq\_doro}\cite{zhai2021doro}: Chisq extension with outlier robustness by ignoring some high-loss samples.
\item \textbf{CVaR\_group}\cite{zhai2021doro}: CVaR-based method minimizing worst-case group risk using known group information.
\item \textbf{Group}\cite{sagawa2019distributionally}: DRO method minimizing worst-case loss across predefined groups.
\item \textbf{Gradient}\cite{namkoong2016stochastic}: DRO method minimizing worst-case loss based on group gradients.
\item \textbf{Variant}: DRO method minimizing worst-case loss based on group variance.
\end{itemize}

To evaluate the performance of our proposed methods, we use the following configurations: \textbf{ERM+NNR}, \textbf{ERM+CMO-KL}, \textbf{ERM+CMO-Chi}, \textbf{ERM+NNR+CMO-KL}, and \textbf{ERM+NNR+CMO-Chi}.

\begin{table}[ht]
\resizebox{0.45\textwidth}{!}{  % 自动调整宽度以适应页面
\begin{tabular}{@{}lc|c|c@{}}
\toprule
Method & \multicolumn{1}{c}{Assay}& \multicolumn{1}{c}{Scaffold}& \multicolumn{1}{c}{Size}  \\ \midrule
& \multicolumn{1}{c}{Class 1(Max\%)}& \multicolumn{1}{c}{Class 1(Max\%)} & \multicolumn{1}{c}{Class 1(Max\%)} \\
\midrule
ERM & 20.3 & 9.3 & 23.5 \\
        ERM+NNR & 20.3 & 14.9 & 24.2 \\
        ERM+CMO-KL & 33.8 & \textcolor{yellow}{\underline{\underline{31.5}}} & \textcolor{yellow}{\underline{\underline{38.3}}} \\
        ERM+CMO-Chi & \textcolor{orange}{\dotuline{41.6}} & 22.1 & 38.3 \\
        ERM+NNR+CMO-KL & \textcolor{yellow}{\underline{\underline{38.0}}} & \textcolor{orange}{\dotuline{31.8}} & \textcolor{orange}{\dotuline{42.2}} \\
        ERM+NNR+CMO-Chi & \textcolor{blue}{\underline{47.2}} & \textcolor{blue}{\underline{34.9}} & \textcolor{blue}{\underline{48.0}} \\
        CVaR & 22.3 & 24.9 & 25.6 \\
        Chisq & 18.0 & 22.1 & 27.4 \\
        CVaR\_doro & 15.1 & 10.0 & 20.2 \\
        Chisq\_doro & 17.4 & 21.1 & 22.7 \\
        CVaR\_group & 19.0 & 14.9 & 34.3 \\
        Variant & 22.3 & 17.3 & 28.2 \\
        Group & 24.6 & 24.6 & 24.5 \\
        Gradient & 15.4 & 10.4 & 33.2 \\
\bottomrule
\end{tabular}
}
\caption{Maximum value of the minority group for drugood\_lbap\_core\_ki (The blue single underline denotes the first place, the orange dotted underline indicates the second place, and the yellow double underline represents the third place.)}
\label{tab:Realdata_MAx}
\end{table}

\subsection{Evaluation Metric}

The primary evaluation metric used in our experiments is \textbf{accuracy}, defined as the proportion of correctly classified samples out of the total number of samples. In addition to the overall accuracy, we focus on the accuracy of the \textbf{minority class}, as well as the overall accuracy, to assess the model's performance in handling imbalanced data and noisy environments.

\subsection{Parameter Settings}

We set the batch size to 32 and the number of epochs to 400. The learning rate is initialized at 0.001. We also perform experiments using random seeds 1, 2, 3, 4, and 5. The pooling method used is mean pooling, and the embedding dimension is set to 32.

% We compared the performance of our methods with baseline approaches on the synthetic dataset with correlation strengths of [0.2, 0.3, 0.4] (~\ref{tab:CS}). Our methods consistently outperform the baseline in improving the minority class performance. However, as the correlation between the invariant and spurious features increases, the performance of our methods declines more significantly compared to the baseline. This is likely due to the reliance on mean representations, which make the spurious features more influential in the overall graph representation as their correlation grows.

% The NNR method shows considerable improvements when the correlation is lower. For instance, \textbf{ERM+NNR} achieves an average of 71.1\%$\pm$1.0\% and a maximum of 83.2, while \textbf{ERM+NNR+CMO-KL} achieves 75.7\%$\pm$0.3\% with a maximum of 83.6. This suggests that NNR helps distinguish samples that are closely associated with spurious features and reduces their impact.

\begin{figure}[h]
  \centering
  \includegraphics[width=0.8\linewidth]{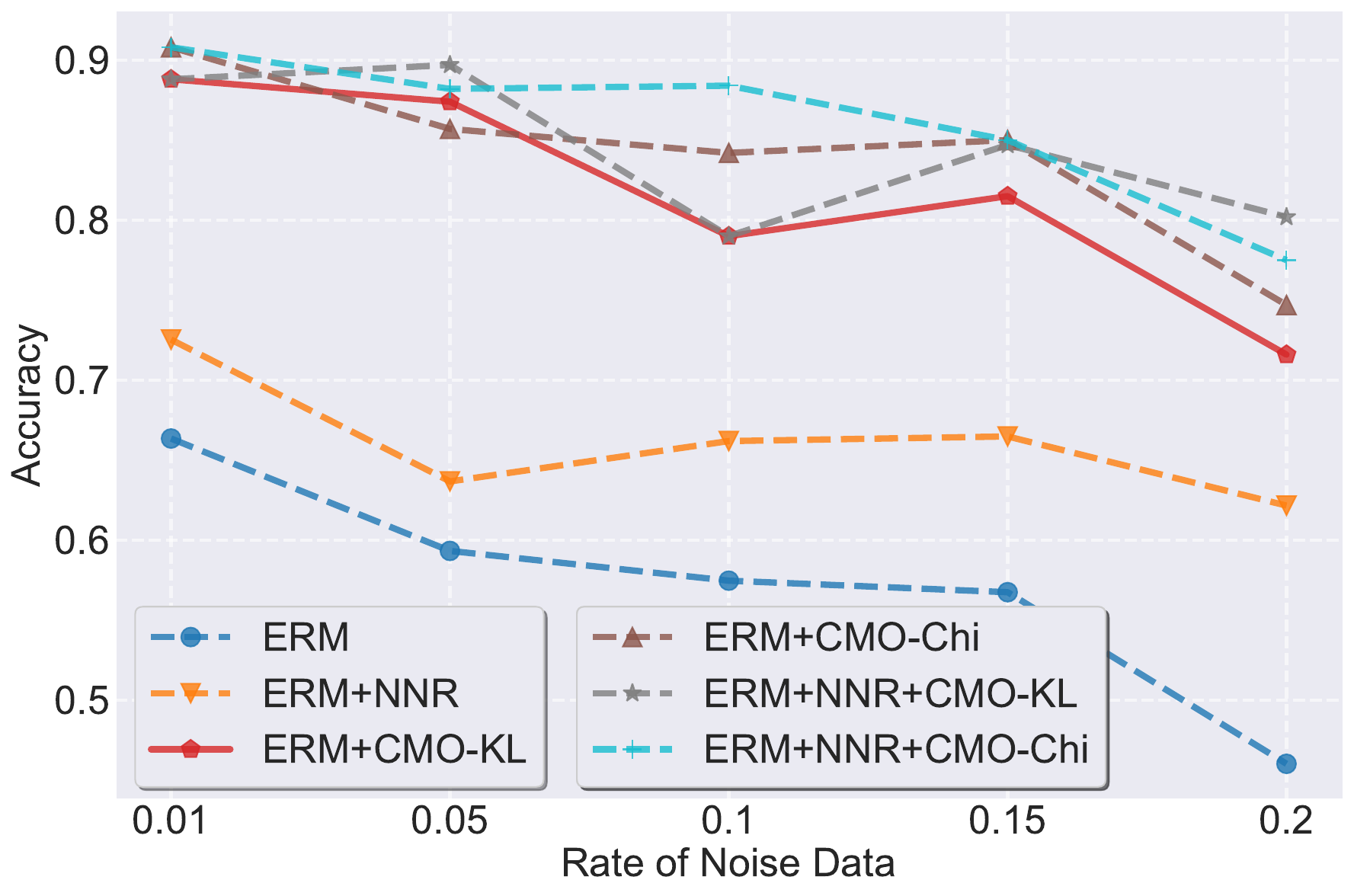}
  \caption{The performance of the minority class with respect to different noise rates on synthetic datasets.}
  \label{fig:noiserate}
\end{figure}

\subsection{Overall Performance}
As shown in ~\ref{tab:NR}, our proposed methods, CMO and NNR, demonstrate clear advantages as the noise ratio increases. CMO consistently improves performance under different noise levels, showing its robustness to class imbalance. NNR further enhances the upper bound of model accuracy, reaching 80.2\% and 79.8\% when combined with CMO-KL and CMO-Chi, which surpass the best baseline results (71.6\% and 74.7\%, respectively). This indicates that NNR effectively reduces the influence of noisy data and helps the model capture more reliable patterns.

On real-world datasets ~\ref{tab:Realdata_Performance1}, our methods consistently achieve superior performance, particularly for the minority class. As shown in ~\ref{tab:Realdata_MAx}, we significantly improve the upper-bound accuracy of minority groups across all tasks, reaching up to 47.2\% in Assay, 34.9\% in Scaffold, and 48.0\% in Size. This demonstrates the strong potential of our approach in handling challenging, imbalanced real-world scenarios. Although we observe increased variance, this is expected due to the absence of dataset-specific weighting strategies during evaluation. Nevertheless, the notable performance gains clearly outweigh this trade-off, confirming the robustness and adaptability of our methods.

However, NNR slightly reduces average accuracy and increases variance (about 0.2\%-0.3\%), suggesting that its noise estimation may occasionally misclassify clean samples, leading to instability. In the Scaffold dataset, NNR alone (\textbf{ERM+NNR}) achieves limited gains (6.9\%$\pm$0.2\%) as the dataset's inherent structure reduces the presence of outliers, limiting the effect of noise suppression.

Overall, these results demonstrate that CMO and NNR provide complementary benefits. CMO offers stable improvements across various settings by addressing class imbalance, while NNR enhances robustness against noise, particularly under challenging conditions with high noise levels. 
% As shown in ~\ref{tab:NR}, CMO and NNR methods outperform others as the noise ratio increases. However, NNR's average performance decreases with higher noise. For example, \textbf{ERM+CMO-KL} achieves 62.8\%$\pm$0.5\%, while \textbf{ERM+NNR+CMO-KL} drops to 59.3\%$\pm$0.8\%. Although NNR reduces average performance, it increases the model's upper bound. \textbf{ERM+NNR} reaches 80.2\% and \textbf{ERM+NNR+CMO-KL} hits 79.8\%, both surpassing baseline models (71.6\% and 74.7\%, respectively). However, this comes at the cost of increased variance (0.2\%-0.3\%), indicating that NNR's current method to noise data selection may cause misclassification and instability.

% On real-world datasets (~\ref{tab:Realdata_Performance1}), our methods remain effective, though the increased variance suggests an enhanced upper bound of performance (~\ref{tab:Realdata_MAx}), but also greater instability. We believe this instability may be attributed to the lack of dataset-specific evaluation weighting.

% On the Scaffold dataset, \textbf{ERM+NNR+CMO-KL} and \textbf{ERM+NNR+CMO-Chi} show notable improvements, but the performance of the NNR method alone (\textbf{ERM+NNR}) is less impressive (6.9\%$\pm$0.2\%). We attribute this to the scaffold dataset's grouping structure, which results in fewer outliers, limiting NNR's effectiveness in this case.
\begin{figure}[t]
  \centering
  \includegraphics[width=0.8\linewidth]{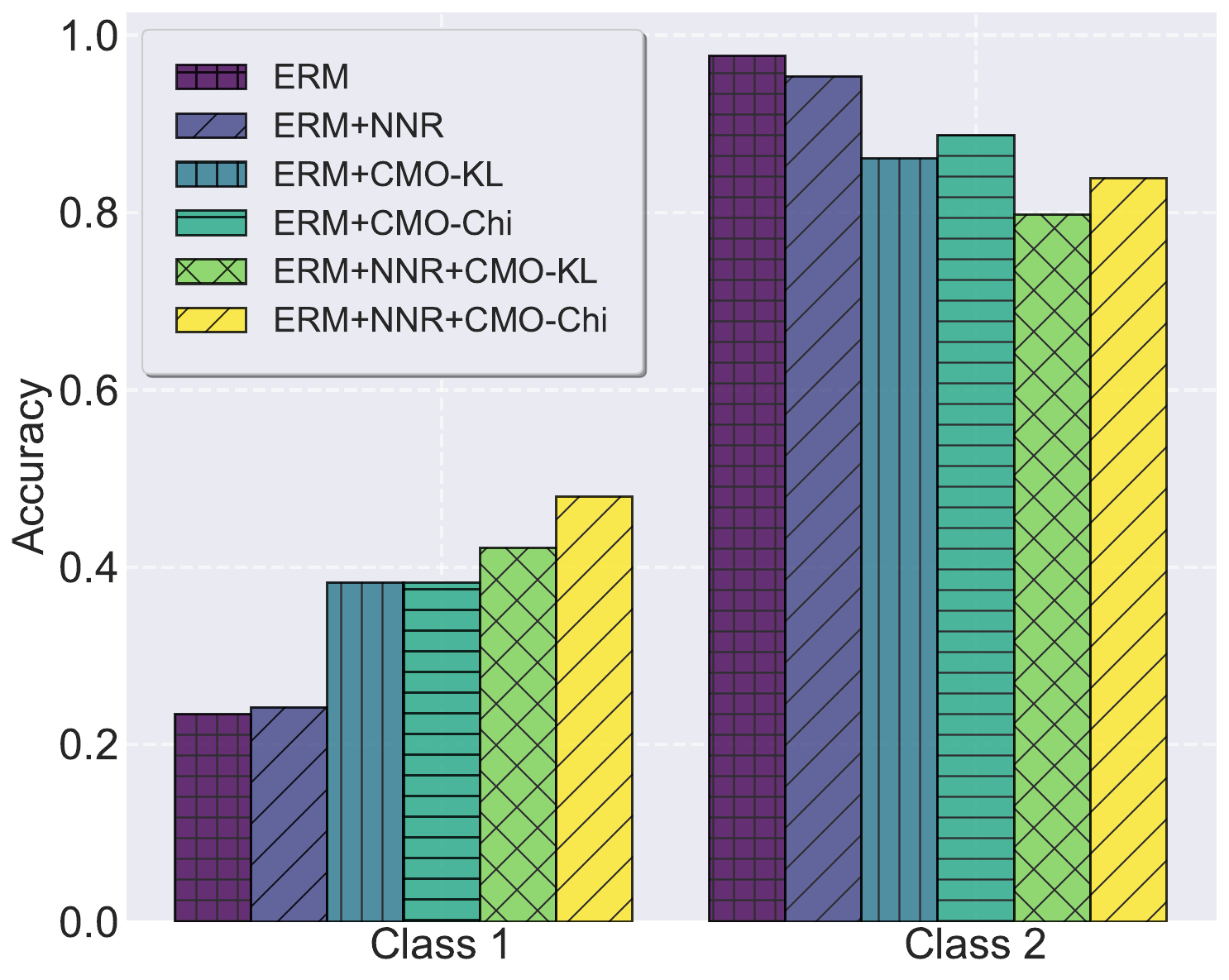}
  \caption{Performance on real-world dataset}
  \label{fig:real-world}
\end{figure}
% \subsection{Ablation Study}
% In this section, we conduct ablation experiments to assess the impact of different components on our model's performance. As shown in ~\ref{tab:CS,tab:NR,tab:Realdata_Performance1,tab:Realdata_MAx,fig:real-world,fig:noiserate}, our full model (NNR + CMO-KL/CMO-Chi) consistently achieves the best results across both synthetic and real-world datasets. This demonstrates the effectiveness of our proposed methods, with NNR mitigating noise and CMO enhancing robustness. Furthermore, as illustrated in ~\ref{fig:Q_value}, our method effectively identifies and prioritizes the worst-performance cases, particularly the minority class. The minority class exhibits a significantly higher Q value than other classes, far exceeding the initial weight baseline (yellow dashed line), highlighting the model's focused optimization on harder cases.
\subsection{Ablation Study}
\begin{figure}[h]
  \centering
  \includegraphics[width=0.8\linewidth]{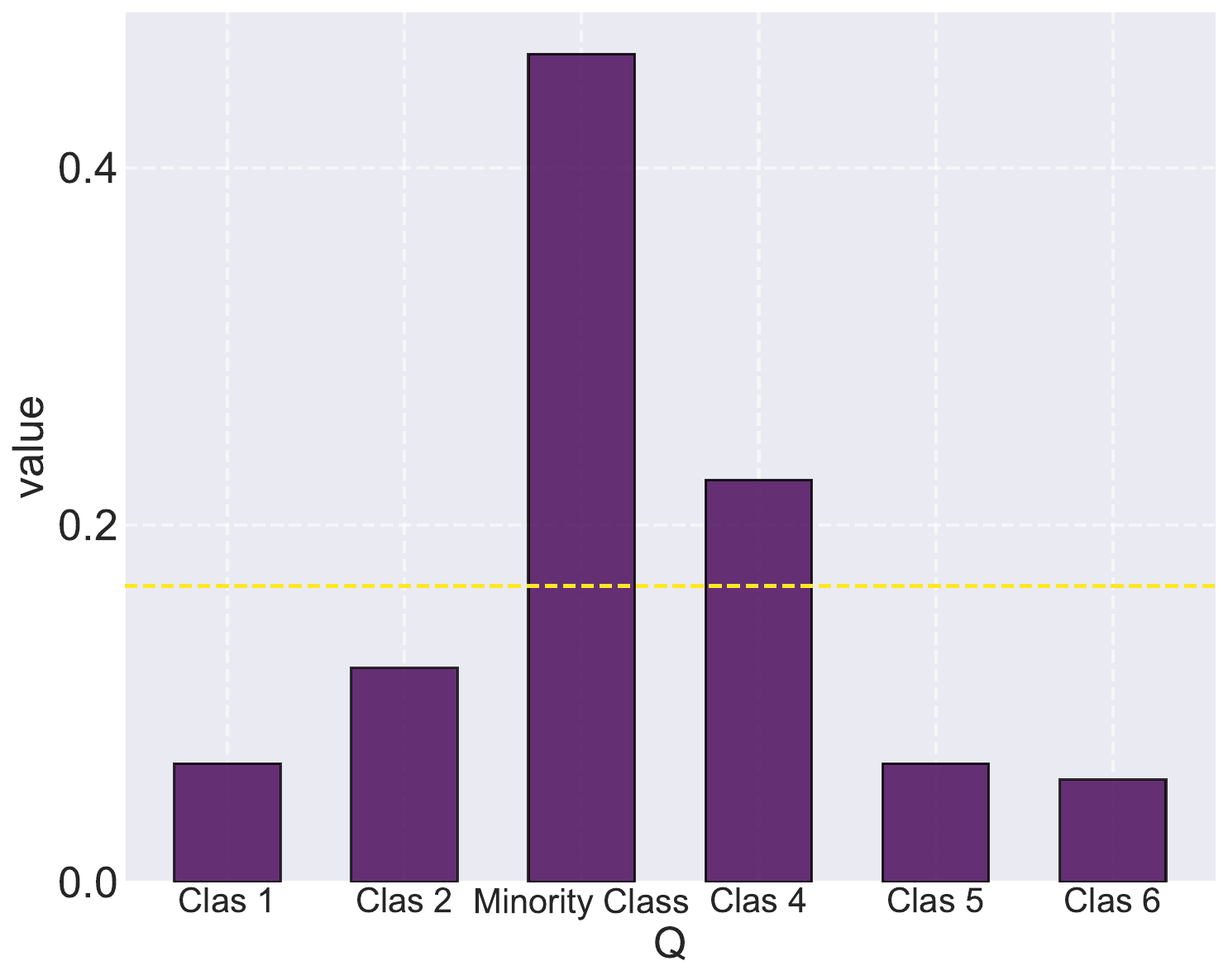}
  \caption{Q value of different classes, yellow line donates the initial weight of each class}
  \label{fig:Q_value}
\end{figure}

In this section, we conduct ablation experiments to assess the impact of different components on our model's performance. As shown in ~\ref{tab:NR}, our full model (NNR + CMO-KL/CMO-Chi) consistently achieves the best results across both synthetic and real-world datasets. This demonstrates the effectiveness of our proposed methods, with NNR mitigating noise and CMO enhancing robustness. Furthermore, as illustrated in ~\ref{fig:Q_value}, our method effectively identifies and prioritizes the worst-performance cases, particularly the minority class. The minority class exhibits a significantly higher Q value than other classes, far exceeding the initial weight baseline (yellow dashed line), highlighting the model's focused optimization on harder cases. \textbf{These results validate the complementary benefits of NNR and CMO, where noise suppression and adaptive reweighting jointly contribute to performance gains.} 
% \vspace{-px}

\subsection{Hyperparameter Sensitive Analysis}
\begin{figure}[ht]
  \centering
  \includegraphics[width=0.8\linewidth]{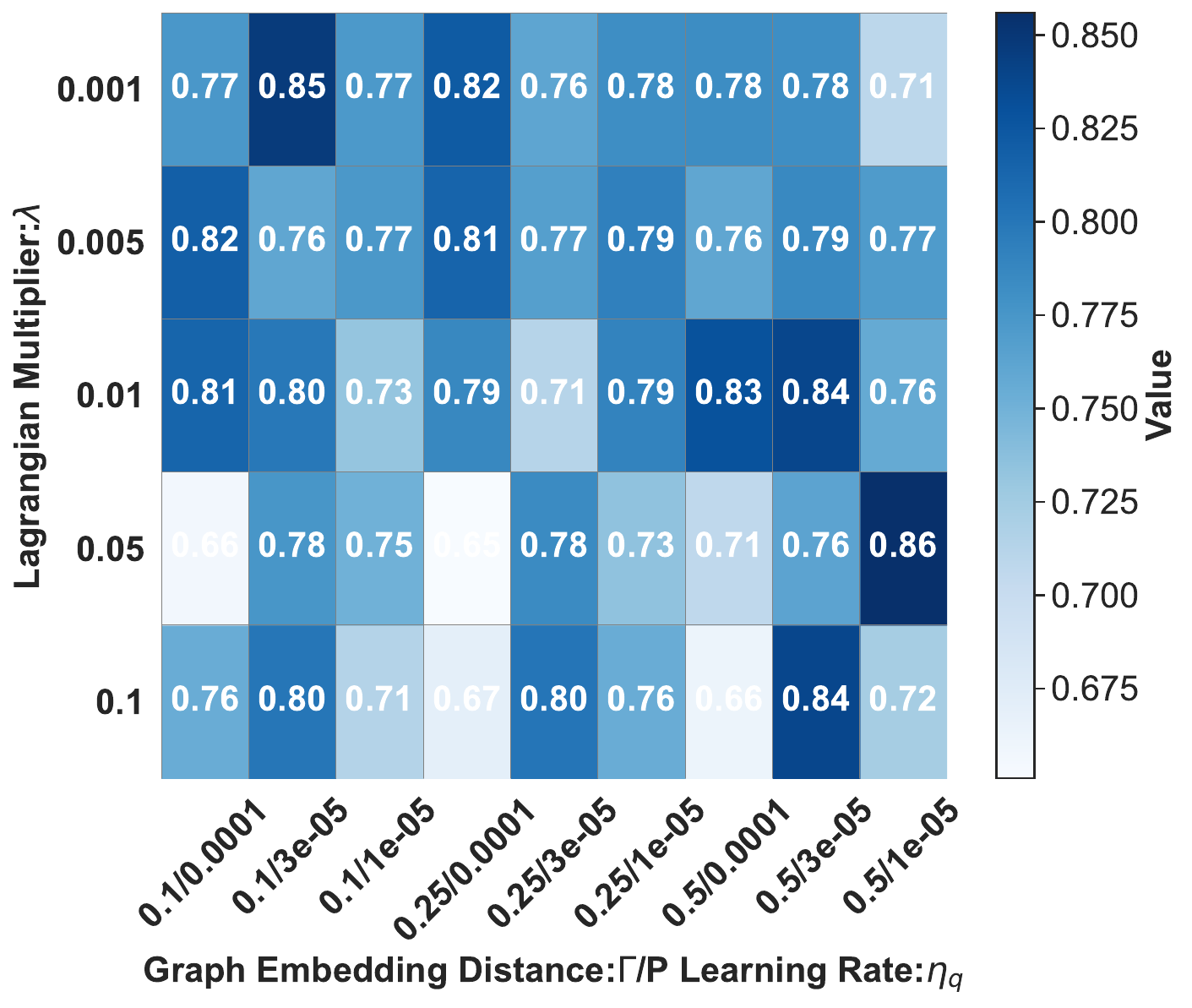}
  \caption{Performance on different hyperparameters}
  \label{fig:hyperparameter}
\end{figure}

To evaluate the influence of key hyperparameters-\textbf{$\Gamma$}, \textbf{$\lambda$}, and \textbf{$\eta_q$}-on the performance of our model, we conduct a sensitivity analysis by varying these hyperparameters within reasonable ranges. Specifically, $\Gamma$ controls the distance metric in NNR, $\lambda$ represent the coefficient of regularization terms in the update formulas (~\ref{eq:Lagrange}), and $\eta_q$ refers to the learning rate for the parameter $q$ during optimization. The experimental results can be seen in ~\ref{fig:hyperparameter}.

\section{Conclusion}
In this work, we address the challenges of class imbalance and noise in Graph OOD learning by proposing CMO and NNR. Experimental results demonstrate that our methods effectively improve the generalization ability and robustness of graph models across diverse environments.

\newpage
\bibliographystyle{ACM-Reference-Format}
% \clearpage
\bibliography{sample-base}
\end{document}